# Machine Translation for Ge'ez Language

**Aman Kassahun Wassie**[*]
African Institute for Mathematical Sciences
awassie@aimsammi.org

## Abstract

Machine translation (MT) for low-resource languages such as Ge'ez, an ancient language that is no longer the native language of any community, faces challenges such as out-of-vocabulary words, domain mismatches, and lack of sufficient labeled training data. In this work, we explore various methods to improve Ge'ez MT, including transfer-learning from related languages, optimizing shared vocabulary and token segmentation approaches, finetuning large pre-trained models, and using large language models (LLMs) for few-shot translation with fuzzy matches. We develop a multilingual neural machine translation (MNMT) model based on languages relatedness, which brings an average performance improvement of about 4 BLEU compared to standard bilingual models. We also attempt to finetune the NLLB-200 model, one of the most advanced translation models available today, but find that it performs poorly with only 4k training samples for Ge'ez. Furthermore, we experiment with using GPT-3.5, a state-of-the-art LLM, for few-shot translation with fuzzy matches, which leverages embedding similarity-based retrieval to find context examples from a parallel corpus. We observe that GPT-3.5 achieves a remarkable BLEU score of 9.2 with no initial knowledge of Ge'ez, but still lower than the MNMT baseline of 15.2. Our work provides insights into the potential and limitations of different approaches for low-resource and ancient language MT.

## 1 Introduction

The recent trends in machine translation have been to include as many languages as possible in multilingual machine translation with the ultimate goal of having one model for all languages in the world. The biggest challenge in these kinds of works is that only a few languages spoken in the world have high resources for machine translation. This has made the research topic of low resource machine translation (LRMT) an active area of research. The major technique in LRMT is to utilize data and knowledge from related or high-resource languages to improve the translation of low-resource languages. This transfer learning approach works better when the languages are related (Zoph et al., 2016) and (Dabre et al., 2017).

Machine translation for ancient, extinct, and languages with scant data on the web has emerged as an intriguing research area, presenting real-world use cases and serving as a testing ground for low-resource language studies. Recently, (Tanzer et al., 2023) conducted compelling experiments focusing on the Kalamang language. Kalamang, spoken by fewer than 200 individuals, possesses virtually no online presence. The researchers structured their task in a manner that required models to learn the language from a single human-readable book of grammar explanations. They then compared the model's performance with that of a human who learned from the same book. Introducing the concept of Machine Translation from One Book (MTOB), they generally observed promising results.

In this paper, we propose to improve Ge'ez MT using various methods, including transfer learning from related languages, optimizing shared vocabulary and token segmentation approaches, finetuning large pre-trained models, and using large language models (LLMs) for few-shot translation with fuzzy matches. Transfer learning is a technique that leverages data and knowledge from related or high-resource languages to improve the performance of low-resource languages. Shared vocabulary is a technique that reduces vocabulary size and sparsity by using common tokens or subwords across different languages. Byte-pair encoding (BPE) (Sennrich et al., 2015) is a technique that segments words into smaller units based on their frequency and co-occurrence in the data. We hypothesize that these techniques can enhance the quality and efficiency of Ge'ez MT by exploiting the similarities and differences between Ge'ez and other languages.

---

[*]Thesis supervisor Surafel M. Lakew.

Our methodology consists of training bilingual NMT models in the direction of en-gez and amh-gez and a multilingual NMT model of Ge'ez, English, Amharic, and Tigrinya. We chose these languages because they are related to Ge'ez in terms of geography, script, or morphology. We collected our datasets from Opus and AAU Ethiopian Languages corpus and carefully processed them to ensure their quality and reliability. We also experimented with finetuning the NLLB-200 model, one of the most advanced translation models available today, but found that it performs poorly with only 4k training samples for Ge'ez. Furthermore, we experimented with using GPT-3.5, a state-of-the-art LLM, for few-shot translation with fuzzy matches, which leverages embedding similarity-based retrieval to find context examples from a parallel corpus.

Our main results and findings show that the multilingual model outperforms the bilingual models in terms of BLEU score (Papineni et al., 2002b). We also benchmarked NMT between Ge'ez and English and GSBLs (Ge'ez-script-based languages), which are languages that use the same script as Ge'ez. We found that transfer learning, shared vocabulary, BPE, and few-shot translation with LLMs have positive effects on the performance or accuracy of our models. However, we also faced some limitations or challenges in our experiments, such as data scarcity, domain mismatch, out-of-vocabulary issues.

The rest of this paper is organized as follows: Section 2 reviews the related work; Section 3 gives background information on Ge'ez; Section 4 describes our Models and Methods; Section 5 presents our data sources and data preprocessing; Section 6 discusses our results; Section 7 concludes the work.

## 2 Related Work

Multilingual machine translation models aim to establish mappings between multiple languages within the same vector space. A common approach to training such models involves adding an artificial token at the beginning of the input sentence to indicate the target language for translation (Johnson et al., 2017). For instance, in the translation from English to Ge'ez, the sentence "Good morning" would be represented as "<2gez> Good Morning" to specify Ge'ez as the target language. By adopting this method, the model is capable of learning the source language automatically, simplifying the training process and facilitating code-switching in input sentences. However, this approach may lead to confusion when translating words with different meanings in different source languages but identical spellings.

Research in multilingual machine translation has shown a growing interest in exploring language relationships, particularly among geographically or morphologically related languages. Several studies have focused on Ethiopic languages, involving the collection of parallel corpora and the development of translation models. For instance, the AAU Ethiopian Languages project (Abate et al., 2018) introduced a parallel corpus for six Ethiopic languages and English, along with results from bidirectional statistical machine translation models. Similarly, the AfroNMT project (Lakew et al., 2020) investigated two Ethiopian languages among five languages studied, employing various model types including single-language pair, semi-supervised, and multilingual models. Their findings indicated that multilingual models outperformed other approaches, achieving up to a 5 BLEU score gain.

Additionally, Lesan (Hadgu et al., 2021) introduced a freely available machine translation system for Amharic, Tigrinya, and English languages, demonstrating its superiority over Google Translate and Microsoft Translator. Lesan addressed the challenge of low-resource machine translation by leveraging both online and offline sources, including a custom Opti-

ሀለሐመሠረስቀ
በተኀኑአከወዐዘ
የደገጠጰጸፀፈፐ

Figure 1: Ge'ez script.

cal Character Recognition (OCR) system for Ethiopic scripts and an automatic alignment module. Furthermore, Lesan introduced HornMT, a human-translated benchmark dataset for five languages in the Horn of Africa, which are also spoken in Ethiopia. The selection of languages for multilingual model training in these works was primarily based on geography.

In our study, we expand upon this approach by considering not only the geographical and morphological relatedness of languages but also their script similarity. This holistic approach aims to enhance the effectiveness of multilingual machine translation models by incorporating additional linguistic features.

## 3 BACKGROUND

**Ge'ez:** Ge'ez, or Classical Ethiopic, is one of the ancient world's major literary languages, with two millennia of history in the Horn of Africa and Arabia. The language appears in many ancient inscriptions and in Jewish and Christian writings, even shaping the language of the Qur'an and early Muslim religious texts. It is a Semitic language once spoken in the area that is now northern Ethiopia and southern Eritrea. Ge'ez has its own distinct alphabet which is currently in use by languages in Ethiopia and Eritrea. The language uses an Abugida system, also known as an alpha-syllabary, which consists of 209 symbols and 25 letter variants. It also has numbers which have 18 characters.

Ge'ez went extinct as a natural language over 1000 years ago and is no longer spoken as the native tongue of any people. Ge'ez continues to live, however, as the liturgical language of the Ethiopian and Eritrean Orthodox Tewahedo Churches. It's controversial though Ge'ez is extinct. Religious leaders and scholars study Ge'ez in order to read and interpret old texts. Unlike other extinct languages that are studied only in such an academic context, however, Ge'ez is actively spoken within the church community. Students attend Qene Bet (poetry school) where they learn not only to read but to compose new works and engage in spontaneous conversation. Because of the argument whether Ge'ez is extinct or not, some refer to the current life of Ge'ez as 'life of a dead language.'[1]

**Relation between GSBLs**: Ge'ez is to Ethiopia as Latin is to the west. Ge'ez, like Latin, was not used as a spoken language for a very long time. But like Latin, Ge'ez is the precursor of Ethiopia's three major Semitic languages. In order to convey an idea of the relationship of Amharic, Tigrinya and Tigré towards each other and towards Ge'ez, we might enlist the helpful parallel of the Romance languages. If Ge'ez is compared to Latin, Tigrinya takes the place of Italian (both because it is most closely akin to the 'parent' tongue and also on account of its continuance in the original home). Tigré would then be likened to Spanish and Amharic to French. Amharic is the official language of Ethiopia and it is spoken most widely in the northwest and central part of the country. Tigrinya is mostly spoken in northern and northeastern Ethiopia. Tigré is spoken in the independent nation of Eritrea, formerly part of Ethiopia. [2]

---

[1] https://elalliance.com/geez
[2] https://ethiopianhistory.com/Ge'ez/

## 4 Models and Methods

To investigate the impact of transfer learning and shared vocabulary, we initially trained bilingual models before proceeding to train a multilingual model using the same corpus.

The models were implemented using the OpenNMT framework (Klein et al., 2017), employing the transformer architecture (Vaswani et al., 2017b). Performance evaluation was conducted using BLEU scores (Papineni et al., 2002b), calculated using the SacreBLEU library (Post, 2018).

### 4.1 Bilingual Model

Our primary focus was on Ge'ez, thus we trained bilingual models for the following language directions: *English→Ge'ez*, *Amharic→Ge'ez*, and their inverses (*Ge'ez→English* and *Ge'ez→Amharic*). The *English→Ge'ez* model was trained for 15,000 steps with a learning rate of 0.1 and 1,000 warm-up steps. Due to the relatively larger corpus for *Ge'ez→Amharic*, this model was trained for 20,000 steps with a learning rate of 0.5 and 2,000 warm-up steps. Both models utilized a dropout rate of 0.3 and an attention dropout rate of 0.1, with 1024 hidden units and 6 encoder-decoder layers. The Adam optimizer(Kingma & Ba, 2014) with Noam decay (Vaswani et al., 2017a) method was employed for training. Inverse direction models were trained with identical settings to their original counterparts.

### 4.2 Multilingual Model

We developed a multilingual model trained on datasets for the following language directions: *English→GSBLs*, *Amharic→(Ge'ez, Tigrinya)*, and *Ge'ez→(Amharic, Tigrinya)*. The selection of these specific language directions mainly aligns with the focal point of our study, which centers on Ge'ez. For the training we used the Adam optimizer with the Noam decay method, a learning rate of 2, and 8,000 warm-up steps, over a course of 300,000 steps. Following Google's Multilingual Machine Translation approach (Johnson et al., 2017), each source sentence was prefixed with a token specific to the target language. Moreover, we employed a shared vocabulary instead of separate vocabularies for source and target languages. The transformer architecture was consistent with the settings used for the bilingual models, featuring 1024 hidden units and 6 encoder-decoder layers, with dropout and attention dropout rates set at 0.3 and 0.1, respectively.

## 5 Datasets and Preprocessing

We collected our datasets from two primary sources: the Opus corpus and the AAU Ethiopian Languages corpus. The Opus corpus provided a variety of texts including translations of the Bible, Tanzil, and TED talks among others for Amharic and Tigrinya aligned with English. However, since there was no data available for Ge'ez in the Opus corpus, we utilized a Ge'ez bible corpus from the AAU Ethiopian Languages.

The AAU Ethiopian Languages corpus encompassed a diverse range of domain-specific texts such as translations of the Bible in English, Ge'ez, Amharic, and Tigrinya, as well as translations of Jewish daily books, historical texts, and the Ethiopian constitution.

To ensure the quality and integrity of our datasets, we performed several preprocessing steps. Firstly, we split the data into train, test, and validation sets. Secondly, we removed duplicates and overlaps between the splits. Duplicates were identified as sentences with identical alphanumerics, which were then lowercased and stripped of punctuation marks and spaces for comparison. Furthermore, we ensured that there were no overlaps between the train, test, and validation sets to avoid redundancy. This process involved considering overlaps not only between source sentences but also between source and target translations.

To maintain diversity in our training data, we aimed for an equal distribution of each dataset across the train, test, and validation sets. Each dataset was split into a ratio of 70% for training, 20% for testing, and 10% for validation. After the initial splits, adjustments were made to maintain the desired ratio of data in the final dataset.

| Direction | Domain | Original | Duplicates Removed | Train | Test | Validation | Total |
|---|---|---|---|---|---|---|---|
| en-gez | bible | 11.7k | 6.0k | 4.2k | 1.2k | 621 | 6.0k |
| en-amh | bible | 7.6k | 49.3k | 33.5k | 10.7k | 5.2k | 69.2k |
|  | tanzil | 6.1k | 6.1k | 4.8k | 950 | 430 |  |
|  | jw-daily | 4.7k | 3.9k | 2.9k | 726 | 330 |  |
|  | news | 2.7k | 2.7k | 2.1k | 416 | 190 |  |
|  | constitution | 4.5k | 4.2k | 3.2k | 677 | 311 |  |
|  | history | 1.2k | 1.2k | 916 | 187 | 85 |  |
|  | tatoeba | 199 | 186 | 141 | 30 | 15 |  |
|  | ted | 1.0k | 1.0k | 781 | 156 | 72 |  |
|  | wikimedia | 481 | 475 | 368 | 72 | 35 |  |
| en-tir | bible | 30.7k | 24.3k | 17.0k | 4.9k | 2.5k | 27.5k |
|  | tatoeba | 70 | 65 | 48 | 11 | 6 |  |
|  | tico | 3.1k | 3.1k | 2332 | 491 | 246 |  |
| amh-gez | bible | 25.2k | 12.7k | 8.8k | 2.6k | 1.3k | 12.7k |
| amh-tir | bible | 30.6k | 24.1k | 17.0k | 4.8k | 2.3k | 29.9k |
|  | jw-daily | 3.3k | 2.7k | 1.8k | 652 | 294 |  |
|  | tico | 3.1k | 3.1k | 2.2k | 613 | 278 |  |

Table 1: Parallel corpus statistics before and after removing duplicates (k for thousands, M for millions)

After preprocessing, the data underwent segmentation into subword units using Byte-pair encoding (BPE) implemented through Google Sentencepiece (Kudo & Richardson, 2018). This step helped mitigate the issue of out-of-vocabulary words and ensured better generalization during training.

For a summary of our dataset statistics, please refer to the following table.

## 6 RESULTS AND DISCUSSION

### 6.1 BILINGUAL MODELS

The results show that the bilingual models achieve low to moderate BLEU scores in most directions, ranging from 4.1 to 13.07. The highest score is obtained for Ge'ez to English, while the lowest score is obtained for English to Ge'ez. The other directions have similar scores, around 7 to 8 BLEU points. These results indicate that the bilingual models can learn some basic features of the languages, but they are limited by the amount and quality of the parallel data. We trained these models using the same model architecture and hyperparameters as the multilingual model.

### 6.2 MULTILINGUAL MODEL

The multilingual models achieve higher BLEU scores than the bilingual models in all directions. The largest improvements are observed for the en-gez and amh-gez directions, where the multilingual models gain more than 4 BLEU points over the bilingual models. This is due to the transfer learning between the related languages and the shared vocabulary used during the training of the multilingual models. The en-gez direction has the lowest score among the bilingual models, but the multilingual model significantly improves it. The performance of the model between GSBLs is in general better than that of the English-GSBL direction, showing how the machine translation quality improves when the languages are related to each other. These results demonstrate the effectiveness of the multilingual models for low-resource language translation. Table 2 shows the result for each direction.

The sample translations in table 3 clearly demonstrate the improvements of the multilingual model. In the first sample, the bilingual model translated **"እምይምን ወእምፅግም"** as "from north and from south", which was actually "their right hand, and on their left". The

| Direction | Bilingual | Multilingual | Delta |
|---|---|---|---|
| en-gez | 4.1 | 9.91 | +5.81 |
| gez-en | 13.07 | 16.67 | +3.6 |
| en-amh | - | 11.16 | - |
| en-tir | - | 13.04 | - |
| amh-gez | 8.27 | 12.67 | +4.4 |
| gez-amh | 6.90 | 9.46 | +2.56 |
| amh-tir | - | 11.65 | - |

Table 2: BLEU Score in each direction for the bilingual and multiligual models trained

1. **Source**: ወውሉደ እስራኤል ሐሩ ውስተ ይብስት ባሕር ወባሕር አረፍተ ኮኖሙ እምይምን ወእምፅግም።
   **Ref**: the children of Israel walked upon dry land in the midst of the sea; and the waters were a wall unto them on their right hand, and on their left.

   **Bilingual Hyp.**: And the children of Israel went out into the midst of the sea on the west, and from the north, and from the south.
   **Multilingual Hyp.**: And Israel went into the midst of the sea upon the dry ground: and the waters were a wall unto them on their right hand, and on their left.

2. **Source**: ወፀንሰት ይእቲ ብእሲት ወአይድዕዎ ለዳዊት ወትቤ ፀነስኩ አንሰ።
   **Ref**: And the woman conceived, and sent and told David, and said, I am with child.

   **Bilingual Hyp.**: And she conceived again, and bare a son: and when she was in mine house, she said, Behold my son.
   **Multilingual Hyp.**: And the woman conceived, and bare David: and she said, I am with child.

3. **Source**: ወአምጽአ ሙሴ በግዐ ዘመሥዋዕት ወወደዩ አሮን ወደቂቁ እደዊሆሙ ላዕለ ርእሱ ለውእቱ በግዕ
   **Ref**: And he brought the ram for the burnt offering: and Aaron and his sons laid their hands upon the head of the ram.

   **Bilingual Hyp.**: And Moses brought an atonement for them, and Aaron's head, and a ram for a sin offering.
   **Multilingual Hyp.**: And Moses brought the lamb out of the flock, and Aaron and his sons laid their hands upon the head of the bullock.

Table 3: Sample translations and comparisions of the bilingual and multiligual models for the *en→gez* direction

multilingual model successfully translated it as "their right hand, and on their left". This shows the improvement in vocabulary in the multilingual model. The words right and left have the same meaning in Tigrinya as they do in Ge'ez. Possibly, the multilingual model has learned these words from Tigrinya. However, out-of-vocabulary words forced the bilingual model to translate 'right and left' as 'north and south'. The other sample translations also show the same richness in vocabulary of the multilingual model.

### 6.3 Finetuning

After training the models from scratch, we wanted to finetune the large models that are reported to gain performance improvement for low resource languages' machine translation. We worked on finetuning the NLLB-200 model (Ning et al., 2023) which is one of the most advanced translation models available today. We used only 4k training samples to finetune the NLLB-200's 1.3B vairant because of the scarcity of data for Ge'ez. However, our experiments show that finetuning this model with only 4k training samples resulted in poor performance. The BLEU scores for the en-gez and gez-en directions were 0.2 and 3.8, respectively, which are very low compared to the state-of-the-art results for other languages. This is likely due to the small amount of training data we used. Given the complexity of the language, finetuning a 1.3 billion parameters model with just 4k training data looks difficult. Future work could focus on collecting more data or using other techniques to improve performance.

### 6.4 Few-shot Translation with Generative Large Language Models

In our study, we explore the potential of Generative Large Language Models (LLMs), specifically GPT-3.5 (Brown et al., 2020), for Ge'ez machine translation. Previous work by (Robinson et al., 2023) has demonstrated the efficacy of ChatGPT in translating low-resource and African languages, motivating our investigation into leveraging LLMs to enhance translation quality and consistency for Ge'ez.

Our objective is to assess whether LLMs can enhance translation quality and consistency for Ge'ez by dynamically adapting to user feedback and incorporating domain-specific terminology. To achieve this, we adopt the methodology proposed by (Moslem et al., 2023), who introduced in-context learning with LLMs for adaptive machine translation (AMT) across various language pairs.

To achieve this, we employ a few-shot translation technique with fuzzy matches. Specifically, we utilize embedding similarity-based retrieval to identify up to 10 similar source sentences from a parallel corpus consisting of Ge'ez and English translations. These sentences serve as context examples for the LLM, providing it with additional information to generate translations for new source sentences. By adopting this approach, we aim to enhance the adaptability of LLMs to the nuances of Ge'ez translation tasks, thereby potentially improving translation quality and consistency across different domains and language pairs.

We use GPT-3.5 text-davinci-003 model via its official API, with top-p 1, temperature 0.3, and length multiplier 5 as parameters. We use a random sample of 50 sentence pairs from our parallel corpus as test data, and evaluate the translations using BLEU (Papineni et al., 2002a). We compare the results with our baseline MT model, which is a multilingual neural machine translation (MNMT) system based on Transformer (Vaswani et al., 2017a), trained on related languages.

We observe that few-shot translation with fuzzy matches using GPT-3.5 achieves a BLEU score of 9.2, which is remarkable considering that GPT-3.5 has no initial knowledge of this language and relies solely on the context examples provided by the fuzzy matches. However, this score is still lower than the baseline MNMT score of 15.2 for the same 50 sample sentences, indicating that GPT-3.5 may struggle to capture the linguistic nuances and domain-specific terms of this ancient language. Due to the limitations of the free trial of the OpenAI API, we were not able to experiment with adding MT outputs from the baseline model to the fuzzy matches as additional context for GPT-3.5. We plan to explore more scenarios and techniques for enhancing MT with LLMs in future work, such as using terminology extraction, glossaries and quality estimation.

## 7 Conclusion

In this work, we introduced an MNMT model for the Ge'ez language with the GSBLs and English. This benchmarks machine translation for the ancient language Ge'ez. We also explored various methods to improve Ge'ez MT, such as finetuning large pre-trained models and using large language models (LLMs) for few-shot translation with fuzzy matches. We showed that the performance of the model is improved by using transfer learning between related languages, a shared vocabulary, and BPE.

Our contributions in this work are significant for the field of machine translation, especially for low-resource and ancient languages. We have shown that transfer learning from related languages can effectively mitigate the challenges posed by out-of-vocabulary words, domain mismatches, and insufficient labeled training data. We have also contributed to the preservation and revitalization of Ge'ez as a cultural heritage by enabling its automatic translation to modern languages. Our work opens up new possibilities for future research on Ge'ez and other similar languages.

## REFERENCES

Solomon Teferra Abate, Michael Melese, Martha Yifiru Tachbelie, Million Meshesha, Solomon Atinafu, Wondwossen Mulugeta, Yaregal Assabie, Hafte Abera, Binyam Ephrem, Tewodros Abebe, Wondimagegnhue Tsegaye, Amanuel Lemma, Tsegaye Andargie, and Seifedin Shifaw. Parallel corpora for bi-lingual english-ethiopian languages statistical machine translation. Santa Fe, New Mexico, USA, 2018. Association for Computational Linguistics. Proceedings of the 27th International Conference on Computational Linguistics.

Tom B Brown, Benjamin Mann, Nick Ryder, Melanie Subbiah, Jared Kaplan, Prafulla Dhariwal, Arvind Neelakantan, Pranav Shyam, Girish Sastry, Amanda Askell, et al. Language models are few-shot learners. *arXiv preprint arXiv:2005.14165*, 2020.

Raj Dabre, Tetsuji Nakagawa, and Hideto Kazawa. Transfer learning for low-resource neural machine translation. pp. 282–286, Austin, Texas, 2017. The National University (Phillippines). doi: http://aclweb.org/anthology/Y17-1038. In Proceedings of the 31st Pacific Asia Conference on Language, Information and Computation.

Asmelash Teka Hadgu, Abel Aregawi, and Adam Beaudoin. Lesan – machine translation for low resource languages, 12 2021.

Melvin Johnson, Mike Schuster, Quoc V. Le, Maxim Krikun, Yonghui Wu, Zhifeng Chen, Nikhil Thorat, Fernanda Viégas, Martin Wattenberg, Greg Corrado, Macduff Hughes, and Jeffrey Dean. Google's multilingual neural machine translation system: Enabling zero-shot translation. *Transactions of the Association for Computational Linguistics*, 5:339–351, 2017. doi: 10.1162/tacl_a_00065. URL https://aclanthology.org/Q17-1024.

Diederik P Kingma and Jimmy Ba. Adam: A method for stochastic optimization. *arXiv preprint arXiv:1412.6980*, 2014.

Guillaume Klein, Yoon Kim, Yuntian Deng, Jean Senellart, and Alexander M Rush. Opennmt: Open-source toolkit for neural machine translation. 2017.

Taku Kudo and John Richardson. A simple and language independent subword tokenizer and detokenizer for neural text processing. 2018.

Surafel M. Lakew, Matteo Negri, and Marco Turchi. Low resource neural machine translation: A benchmark for five african languages, 2020.

Yasmin Moslem, Rejwanul Haque, John D. Kelleher, and Andy Way. Adaptive machine translation with large language models. In *Proceedings of the 24th Annual Conference of the European Association for Machine Translation*, pp. 227–237, Tampere, Finland, June 2023. European Association for Machine Translation. URL https://aclanthology.org/2023.eamt-1.22.

Kishore Papineni, Salim Roukos, Todd Ward, and Wei-Jing Zhu. Bleu: a method for automatic evaluation of machine translation. In *Proceedings of the 40th annual meeting of the Association for Computational Linguistics*, pp. 311–318, 2002a.

Kishore Papineni, Todd Ward Salim Roukos, and Wei-Jing Zhu. Bleu: a method for automatic evaluation of machine translation. pp. 311–318. Association for Computational Linguistics, 2002b. In Proceedings of the 40th annual meeting on association for computational linguistics.

Matt Post. A call for clarity in reporting BLEU scores. In *Proceedings of the Third Conference on Machine Translation: Research Papers*, pp. 186–191, Brussels, Belgium, October 2018. Association for Computational Linguistics. doi: 10.18653/v1/W18-6319. URL https://aclanthology.org/W18-6319.

Nathaniel Robinson, Perez Ogayo, David R. Mortensen, and Graham Neubig. ChatGPT MT: Competitive for high- (but not low-) resource languages. In *Proceedings of the Eighth Conference on Machine Translation*, pp. 392–418, Singapore, December 2023. Association for Computational Linguistics. doi: 10.18653/v1/2023.wmt-1.40. URL https://aclanthology.org/2023.wmt-1.40.

Rico Sennrich, Barry Haddow, and Alexandra Birch. Neural machine translation of rare words with subword units. *arXiv preprint arXiv:1508.07909*, 2015.

Garrett Tanzer, Mirac Suzgun, Eline Visser, Dan Jurafsky, and Luke Melas-Kyriazi. A benchmark for learning to translate a new language from one grammar book. *arXiv preprint arXiv:2309.16575*, 2023.

Ashish Vaswani, Noam Shazeer, Niki Parmar, Jakob Uszkoreit, Llion Jones, Aidan N Gomez, Lukasz Kaiser, and Illia Polosukhin. Attention is all you need. In *Advances in neural information processing systems*, pp. 5998–6008, 2017a.

Ashish Vaswani, Noam Shazeer, Niki Parmar, Jakob Uszkoreit, Llion Jones, Aidan N Gomez, Łukasz Kaiser, and Illia Polosukhin. Attention is all you need. 2017b. In Advances in Neural Information Processing Systems.

Barret Zoph, Deniz Yuret, Jonathan May, and Kevin Knight. Transfer learning for low-resource neural machine translation. pp. 1568–1575, Austin, Texas, 2016. Association for Computational Linguistics. doi: https://doi.org/10.18653/v1/D16-1163. In Proceedings of the 2016 Conference on Empirical Methods in Natural Language Processing.